# FISH RECOGNITION BASED ON THE COMBINATION BETWEEN ROBUST FEATURES SELECTION, IMAGE SEGMENTATION AND GEOMETRICAL PARAMETERS TECHNIQUES USING ARTIFICIAL NEURAL NETWORK AND DECISION TREE

**Mutasem Khalil Sari Alsmadi[1], Prof.Dr Khairuddin Bin Omar[2] , Prof.Dr.Shahrul Azman Noah[3] and Ibrahim Almarashdah[4]**

**Faculty of Computer Sciences**
**University Kebangsaan Malaysia**

*Abstract---* We presents in this paper a novel fish classification methodology based on a combination between robust feature selection ,image segmentation and geometrical parameter techniques using Artificial Neural Network and Decision Tree. Unlike existing works for fish classification, which propose descriptors and do not analyze their individual impacts in the whole classification task and do not make the combination between the feature selection, image segmentation and geometrical parameter, we propose a general set of features extraction using robust feature selection, image segmentation and geometrical parameter and their correspondent weights that should be used as a priori information by the classifier. In this sense, instead of studying techniques for improving the classifiers structure itself, we consider it as a "black box" and focus our research in the determination of which input information must bring a robust fish discrimination. The study area selected for our proposed method from global information system (GIS) on Fishes (fish-base) and department of fisheries Malaysia ministry of agricultural and Agro-based industry in putrajaya, Malaysia region currently, the database contains 1513 of fish images. Data acquired on 22th August, 2008, is used. The classification problem involved the identification of 1513 types of image fishes; family ,Scientific Name , English name , local name, Habitat , poison fish and non-poison .The main contribution of this paper is enhancement recognize and classify fishes based on digital image and To develop and implement a novel fish recognition prototype using global feature extraction, image segmentation and geometrical parameters, it have the ability to Categorize the given fish into its cluster and Categorize the clustered fish into poison or non-poison fish, and categorizes the non-poison fish into its family . Both classification and recognition are based on combination between robust feature selection, image segmentation and geometrical parameter techniques.

## I. INTRODUCTION

Recognition and cataloging are the vital facets in this up-to-the-minute era of research & development, hence exploiting the accessible techniques in Artificial Intelligence (AI) and Data Mining (DM) to achieve optimal production levels, examination procedures, and enhancing methodologies in most fields principally in the agricultural domain.

Artificial neural networks are defined as computational models of nervous system. Significantly natural organisms do not only possess nervous system; in fact they also evolve genetic information stored in the nucleus of their cells (genotype). Furthermore, the nervous system as a whole is part of the phenotype which is derived from the genotype through a specific development process. The information specified in the genotype determines assorted aspects of the nervous system which are expressed as innate behavioral tendencies and predispositions to learn (Parisi, 2002), acknowledges that when neural networks are viewed in the broader biological context of Artificial Life, they tend to be accompanied by genotypes and to become members of budding populations of networks in which genotypes are inherited from parents to offspring. Many researchers such as Holland, Schwefel, and Koza, have stated that Artificial Neural Networks are evolved by the utilization of evolutionary algorithms.

Moreover, there are several methods that can make the computer more intelligent and to give it enough intelligence to recognize and to understand the images that the user gives to it. One of this ways is using the Artificial Intelligence (AI) and Decision Tree (DT) Science. Using one of AI techniques such as Neural Network (NN) will help us in recognize and then classify the entered image which will give a big contribution in the agriculture domain especially in fish recognition and classification.

## II. PROBLEM STATEMENT

Several efforts have been devoted to the recognition of digital image but so far it is still an unresolved problem. ( Bai el al,.2008 Kim and Hong ,2009), due to distortion, noise, segmentation errors, overlap, and occlusion of objects in color images. Recognition and classification as a technique gained a lot of attention in the last years wherever many scientists utilize these techniques in order to enhance the scientific fields. Fish recognition and classification still active area in the agriculture domain and considered as a potential research in utilizing the existing technology for encouraging and pushing the agriculture researches a head. Although advancements have been made in the areas of developing real time data collection and on improving range resolutions (Patrick et al,. 1992 and Nery et al. 2006), existing systems are still limited in their ability to detect or classify fish. And despite the widespread development in the world of computers and software. There are many of people die every day because they do not have the ability to distinguish between poison fish and non- poison.

Recognition ability from image can also be applied into computer system for automated recognition based on not just the text input but also the shape of images. The recognition of patterns (fishes) from scanned images of documents has been a problem that has received much attention in the fields of image processing, pattern recognition and artificial intelligence. Classical methods such as experiments on species pattern recognition (plants, rats, and many more) and particularly fishes do not suffice for the recognition of their shape due to some reasons. Firstly; considering that the image is to be run into a neural network system, the recognition of that image is subjected to the disturbance or spoilage of the system due to some characteristics such noise and climate. Secondly; related to the recognition of the fish shape is made difficult because there are no hard-and-fast rules of the ability of recognizing and defining the appearance of a given fish.

The Object classification problem lies at the core of the task of estimating the prevalence of each fish species. They mentioned about that this issue still has a problem with classification and identification of fish species, and the authors understand that any solution to the automatic classification of the fish should address the following issues as appropriate:

1) *Arbitrary fish size and orientation*; fish size and orientation are unknown a priori and can be totally arbitrary;

2) *Feature variability*; some features may present large differences among different fish species;

3) *Environmental changes*; variations in illumination parameters, such as power and color and water characteristics, such as turbidity, temperature, not uncommon. The environment can be either outdoor or indoor;

4) *Poor image quality*; image acquisition process can be affected by noise from various sources as well as by distortions and aberrations in the optical system;

5) *Segmentation failures*; due to its inherent difficulty, segmentation may become unreliable or fail completely; And the vast majority of research-based classification of fish points out that the basic problem in the classification of fish; they typically use small groups of features without previous thorough analysis of the individual impacts of each factor in the classification accuracy (Al smadi ,et al 2009 ; Nery, et al, 2006 and Lee, et al,2004).

### III.     MATERIALS AND METHODS

The study area selected for our proposed method from global information system (GIS) on Fishes (fish-base) and department of fisheries Malaysia ministry of agricultural and Agro-based industry in putrajaya, Malaysia region currently, the database contains 2000 of fish images. Data acquired on 22th August, 2008, is used. The classification problem involved the identification of 1500 types of image fishes; family ,Scientific Name , English name , local name, Habitat , poison fish and non-poison, based on set of extraction feature .The main contribution of this paper is  enhancement recognize and classify fishes based on digital image. Both classification and recognition are based on combination between robust feature selection, image segmentation and geometrical parameter techniques.Based on the problem's requirements, a model of solving the issue that has been proposed is going to be set, which is Decision Tree (DT) and multilayer-perceptron (MLP) model using the back-propagation (BP) algorithm. BP is one of the well-known neural network classifiers. In order to utilize the ability of multilayer-perceptron model with BP algorithm and DT in increasing and consolidating the recognition and classification results. This can be done by assigning the inputs of the fish image to image analysis for the feature extraction. At this moment the extracted features are going to be utilized by the MLP model using BP algorithm. In order to recognize and classify the fish based on its image attributes comes from feature extraction and analyzing the fish image, the output of the MLP model with BP algorithm will be entered into the DT. In the end, the DT will categorize the given fish's output into its cluster, therefore; categorizing the clustered fish into poison fish and non-poison fish. And categorizes the non-poison fish into its family.

## IV. ARTIFICIAL NEURAL NETWORK CLASSIFICATION

An artificial neural network is a mathematical model to processing of information. It is an approximate reply of the human brain behaviour across the emulation of the operations and biological neurons connectivity (Tsokalas and Uhring, 1997; Carvajal ,et al 2006).

Successful adaptation of artificial neural network to remote sensing is mainly due to:

Its efficiency is high because it does not need to take assumptions about distribution functions models

(Benediktsson et al. 1990, 1993; Schalkoff, 1992; Carvajal ,et al ,2006).

Time computation use to be shorter than other methods (Bankert 1994; Cote and Tatnall 1995; Carvajal ,et al 2006).

It allows incorporating a priori knowledge about class objective and real limits (Brown and Harris 1994;

Foody 1995a, b; Carvajal ,et al ,2006).

It allows to management of multi-source spatial data, getting synergy classification results (Benediktsson et al. 1993; Benediktsson and Sveinsson 1997; Carvajal ,et al ,2006).

## V. NEURAL NETWORK ARCHITECTURE

Like human nervous system, an artificial neural network consists of a set of interconnected nodes, called neurons (see Figure 1). Its outlet depends on weighted inlet information from all inlet nodes (Atkinson and Tatnal, 1997; Carvajal ,et al ,2006).

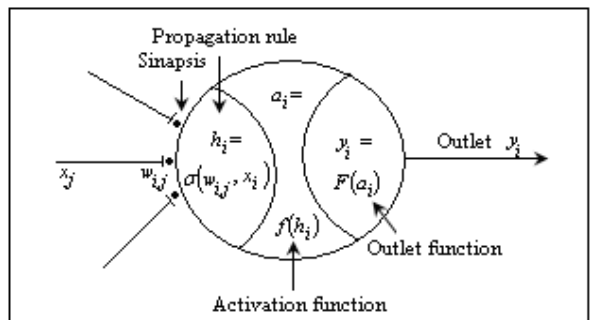
Figure 1. Artificial neuron schema

Neurons use to be clustered in functional units or levels. Figure 2 shows neural network design used in this work. It is composed of tree levels in which inlet level neurons are four channels in multi-spectral satellite image; outlet level neurons are detected classes and hidden level is composed of operational neurons.

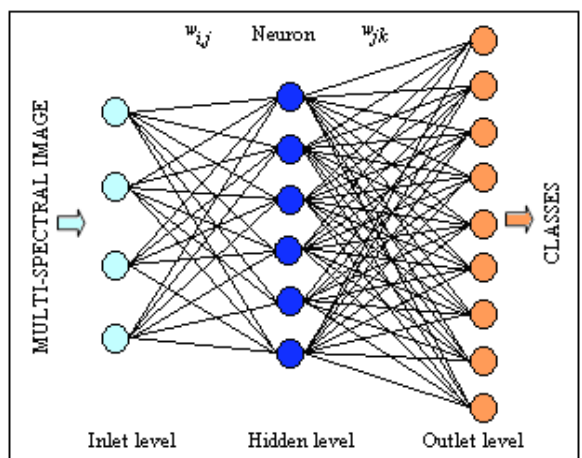
Figure 2. Artificial Neural Network structure

An increment of hidden levels in the design could allow more complex problems resolution, but its generalization ability decreases and training time increments (Foody, 1995b; Carvajal ,et al ,2006). (Lippmann ,1987; Carvajal ,et al ,2006) suggests that if one aggregates a second hidden level, its maximum number of nodes must be limited to triple number of nodes in the first level. In this work, we added one neuron to inlet level, corresponding to texture analysis when it was performed. Response of each neuron depends on function activation evaluation (Figure 3).

Back-propagation method was used to train our neural network. This is, information travels forward and backward in training process.

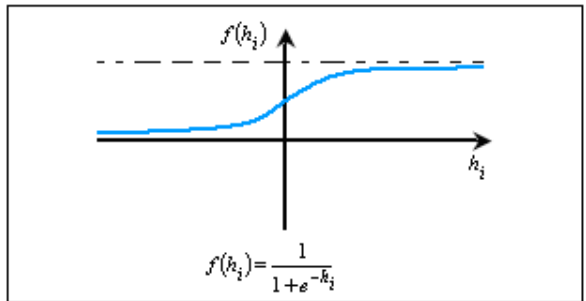

Figure 3. Activation function

## VI. TRAINING PROCESS : BACK-PROPAGATION

In training process neural network adjusts its free parameters (Yao, 1995; Carvajal ,et al ,2006). It starts with a set of initial weights associated to relationship between pairs of neurons (or synapses). These values change depending of error committed in classification, following Generalized Delta Rule (Pao, 1989; Carvajal ,et al ,2006). This process is repeated iteratively until convergence is reached in two phases. Firstly a set (xp) of inlet data is introduced in neural network. This set is propagated forward network, delivering an outlet (yp), which is compared with desired outlet (dp), obtained from training sites. Error committed in classification at this moment in training process (ep) is calculated according to (1):

$$e_p = \frac{1}{2}\sum_{k=1}^{M}\left(d_{pk} - y_{pk}\right)^2 \qquad (1)$$

Where k is neurons outlet index in last level, and M is total number neurons in this level. Secondly, classification error is propagated backward, modifying weight factors wp using Rumerhart rule (2) (Atkinson and Tatnal, 1997; Carvajal ,et al ,2006).

$$\Delta w_{ji}(t+1) = \eta \frac{\partial e_p}{\partial w_{ji}} + \alpha \Delta w_{ji}(t) \qquad (2)$$

Where η is learning rate and α is momentum factor. The iterative process finishes when difference between classification error in two iterations t and t+1 in (2), falls under a threshold value. Parameters affect to velocity which convergence is reached. So, if learning rate is high, mathematical solution will be early encountered but it is possible that in next iteration, classification increases. Using a high momentum factor, these oscillations can be reduced.

## VII. THE FEATURE EXTRACTION APPROACH

As a first step we have set out to determined a largest set of features. For each fish species, we have computed 47 different features, which can be divided in four main groups, differentiated by the type of extracted information. Those groups and their corresponding numbers of features are:

### A. Preprocessing :

In the pre processing step, an image filtration is required to remove noise from the image, through background unification process to ease the isolation of patterns of interest (fish) from the background of the image in the next step. Also, the image might need adjustment of its rotation. The input of a pattern recognition system is typically a digital image. The digital images were downloaded to a personal computer having a Pentium 200MMX microprocessor and 96 MB of RAM .

### B. Image Processing:

Image processing involved in isolation of patterns of interest (fish) from the background of the image; and color extraction of a fish image via space RGB and dividing the neighboring pixel in term of color similarity into a number of groups, and finds the correlation between similar groups. The digital images converted from the native Kodak digital camera format (KDC) to the 8-bit colour bitmap format (BMP). The size of the images was 856x804 pixels. After the BMP images were obtained, they were preprocessed with the Image Processing Toolbox v2.0 for MATLAB v8.0.

The BMP images were converted to indexed images based on a red-green-blue (RGB) colour system. Each pixel of an image was classified into one of 256 categories, represented by an integer in the range from 0 (black) to 255 (white).

*C. Segmentation :*

Segmentation is one of the deepest problems in pattern recognition. The method involves analysis of each picture to find the contours of the pattern. Dividing the fish into a certain number of segments, measurements and characteristics (size, shape, color, geometrical parameter), and analyze the color in each segment. The purpose of segmentation is identifying the objects to be recognized into the raw data and storing them into database.

*D. Feature Extraction:*

The goal of a feature extraction determines a largest set of features. In which characterizing the objects to be recognized by measurements whose values are likely very similar for those in the same class, and very different for those in different ones. This leads to the idea of seeking for distinguishing features that are invariant to irrelevant transformations of the input. For example, the absolute position of an object identified in the acquired scene is irrelevant to the category of that object and thus the representation to be used should be insensitive to its absolute position.

The main categories of feature extraction approach can be listed below:

➢ *Size measurements*

➢ *Shape and Texture measurements*

➢ *Color signatures*

➢ *Geometrical Parameters*

The following two figures are illustrate the fish feature extraction based on the four categories of feature extraction mentioned above.

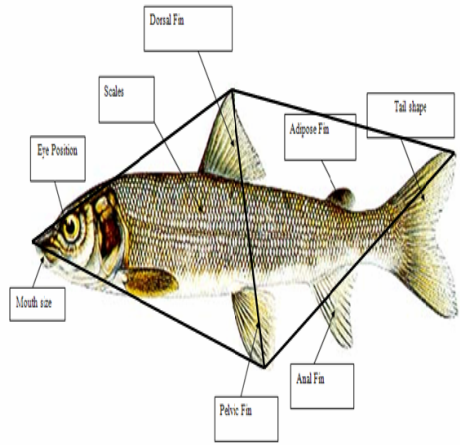

Figure 5a: Fish External Anatomy (Fish Features)

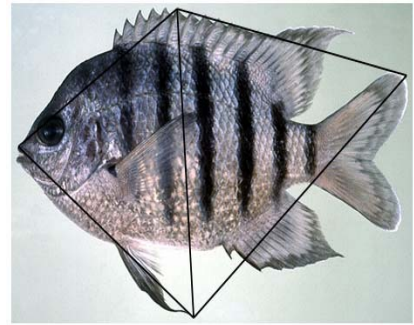

Figure 5b: Fish External Anatomy (Fish Features)

Figures 5a and 5b shows the fish external anatomy that is chosen to be used in fish classification. As follows, the explanation of each category illustrated in both figures:

➢ *Size Measurements:*

This group of measurements consists of planar measurements on the fish's area, and fish's length and width. These features are not invariant under translation, scales and rotation; they are fundamentally role in computing other relevant features.

➢ *Shape and Texture Measurements:*

Using shape measurements, the external contour and edge detection of the pattern for each fish and to determine the significant similarity part, such as the tail shape. The geometrical parameters obtained by shape measurements as well. Using texture measurements determine the dorsal, anal, pelvic and adipose fins.

➢ *Color Signature:*

Kim and Hong, (2009) Color and texture are essential features for image segmentation and recognition since these features are commonly observed in most images, especially in color textured images of natural scenes, where natural objects, such as the flowers or the wild animals, have their own color and texture. Keenleyside (1979); the dorsum and ventral colorations constitute very important features that might be used to discriminate different fish species. Based on this fact, the usage of this information by assigning to each fish species a color signature can be beneficial, which is composed of the average color of the dorsum and the ventral region of the fish.

➢ *Geometrical Parameters*

Through the usage of geometrical parameters, the eye position and size of mouth can be determined. Besides dividing the fish into two triangles, which can be a significant step in obtaining a high accuracy of fish classification. According to figures 5a and 5b, a different triangles are drawn based on the maximum and minimum points on the x-axes as well as y-axes, finalizing the triangle drawing process by connecting lines between the maximum and the minimum points on x-axes with the maximum and minimum points on y-axes. This will lead to the classification process through measuring the position of fish's eyes, size of mouth, and the similarity of the triangles, and coordinates of the triangles peaks.

## VIII.  Experiment design
### Table 1

| Scientific Name of fish family | Number of fish type | Training set | Test set |
|---|---|---|---|
| Istiophoridae | 15 | 15 | 10 |
| leiognathidae | 10 | 10 | 10 |
| Acropomaatidae | 15 | 15 | 10 |
| Scombridae | 10 | 10 | 10 |
| Stromateidae | 15 | 15 | 10 |
| Triacanthidae | 10 | 10 | 10 |
| Poison fish | 25 | 25 | 25 |

Our Prototype system has been applied over 7 different fish families, each family has a different number of fish types, Our sample consists of distinct 100 of fish images , 100 images used for trained neural network and 75 used for tested , which are illustrated in table 1. The implementation of our prototype system including the training and testing processes of all fish families resulted that the classification accuracy of a fish type has a high quality and accurate classification result achieving 97.4%.

Regarding this percentage of the obtained result, the significance of the combination of the image segmentation, feature extraction and the geometrical parameters is very high and dependent in order to obtain a high accuracy of a classification result.

## IX.    Conclusion

This study was undertaken to develop an ANN and DT to classify fish  images taken from the global information system (GIS) on Fishes (fish-base) and   department of fisheries Malaysia ministry of agricultural and Agro-based industry in putrajaya, Malaysia region currently, the database contains 1513 of fish images. Data acquired on 22th August, 2008, is used. The classification problem involved the identification of 1513 types of image fishes; family ,Scientific Name , English name , local name, Habitat , poison fish and non-poison .The main contribution of this paper is enhancement recognize and classify fishes based on digital image and To develop and implement a novel fish recognition prototype using global feature extraction, image segmentation and geometrical parameters .

Our prototype system it have the ability to Categorize the given fish into its cluster and Categorize the clustered fish into poison or non-poison fish, and categorizes the non-poison fish into its family . Both classification and recognition are based on combination between robust feature selection, image segmentation and geometrical parameter techniques.

The performance of the ANNs and DT was compared and the success rate for the identification of corn was observed to be as high as 96.4%, the results indicate the potential of ANNs and DT for fast image recognition and classification. Fast image recognition and classification based on the combination between robust feature selection, image segmentation and geometrical parameter techniques using Artificial Neural Network and Decision Tree can be useful in the control of real-world and It will contribute a lot in the agriculture domain in Malaysia and Scientists in the same field, which they can utilize it to do new researches in investigating and exploring fishes and Marine world. Moreover, researchers, students, and amateurs will benefit from it in their own research.

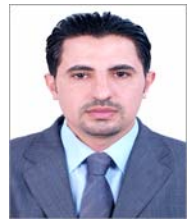
**Mutasem Khalil Sari Al Smadi** received his BS degree in Software engineering in 2006 from Philadelphia University, Jordan, his MSc degree in intelligent system in 2007 from University Utara Malaysia, Malaysia; He has published numbers of paper in IEEE and International Journal of Computer Science and Network Security; currently he is doing PhD in Intelligent System in University Kebangsaan Malaysia in Malaysia.

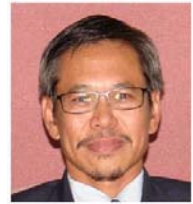
**Khairuddin Omar** is an Associate Professor in Faculty of Information Science and Technology, University Kebangsaan Malaysia, 43600 UKM Bangi, Malaysia. His research interest includes Arabic/Jawi Optical Text Recognition, Artificial Intelligence for Pattern Recognition & Islamic Information System.

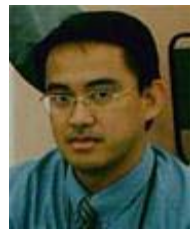
**Shahrul Azman Noah** is currently an associate professor at the Faculty of Information Science and Technology, University Kebangsaan Malaysia (UKM). He received his MSc and PhD in Information Studies from the University of Sheffiled, UK in 1994 and 1998 respectively. His research interests include information retrieval, knowledge representation, and semantic technology. He has published numerous papers related to these areas. He currently leads the knowledge technology research group at ukm

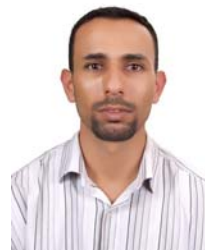
IBRAHEM ALMRASHDEH received his BS degree in physical education in 2001 from University of Jordan, Jordan, his MSc degree in ICT in 2007 from University Utara Malaysia, Malaysia; He has published paper inIEEE.